\documentclass [10pt,twocolumn]{IEEEtran}
\usepackage{algorithm,amsbsy,amsmath,amssymb,epsfig,bbm,mathrsfs,multirow,amsthm}
\usepackage[T1]{fontenc}
\usepackage[latin9]{inputenc}
\usepackage{array,multirow}
\usepackage{color}
\usepackage{cite}
\usepackage{float}
\usepackage{makecell}
\usepackage[x11names,dvipsnames,table]{xcolor} 
\usepackage{colortbl} 
\usepackage{multirow}
\usepackage{subcaption} 
\usepackage{lipsum}
\usepackage{booktabs}
\definecolor{mygray}{gray}{0.6}
\definecolor{myblue}{rgb}{0.8,0.85,1} 
\usepackage{array}
\newcolumntype{L}[1]{>{\raggedright\let\newline\\\arraybackslash\hspace{0pt}}m{#1}}
\newcolumntype{C}[1]{>{\centering\let\newline\\\arraybackslash\hspace{0pt}}m{#1}}
\newcolumntype{R}[1]{>{\raggedleft\let\newline\\\arraybackslash\hspace{0pt}}m{#1}}

\usepackage{array, tabularx, boldline}
\usepackage{eurosym}
\usepackage{amstext} 
\DeclareRobustCommand{\officialeuro}{%
  \ifmmode\expandafter\text\fi
  {\fontencoding{U}\fontfamily{eurosym}\selectfont e}}

\usepackage{epstopdf}
\usepackage{supertabular}
\usepackage[noend]{algpseudocode}  
\usepackage{mathtools}

\usepackage{array, tabularx, boldline}
\usepackage{graphicx}
\usepackage{cellspace}
\newcolumntype{b}{X}
\newcolumntype{s}{>{\hsize=.5\hsize}X}

\begin{document}
\title{\huge Cognitive Radio Network Throughput Maximization with Deep Reinforcement Learning}
\author{Kevin Shen Hoong Ong, Yang Zhang, Dusit Niyato \\
\IEEEauthorblockA{\textit{School of Computer Science and Engineering} \\
\textit{Nanyang Technological University Singapore}\\
ongs0129@ntu.edu.sg, yzhang28@ntu.edu.sg, dniyato@ntu.edu.sg}
}

\maketitle
\begin{abstract}
Radio Frequency powered Cognitive Radio Networks (RF-CRN) are likely to be the eyes and ears of upcoming modern networks such as Internet of Things (IoT), requiring increased decentralization and autonomous operation. To be considered autonomous, the RF-powered network entities need to make decisions locally to maximize the network throughput under the uncertainty of any network environment. However, in complex and large-scale networks, the state and action spaces are usually large, and existing Tabular Reinforcement Learning technique is unable to find the optimal state-action policy quickly. In this paper, deep reinforcement learning is proposed to overcome the mentioned shortcomings and allow a wireless gateway to derive an optimal policy to maximize network throughput. When benchmarked against advanced DQN techniques, our proposed DQN configuration offers performance speedup of up to 1.8x with good overall performance.
%
\end{abstract}
\section{Introduction and Related Work}
Internet of Things (IoT) enables large amounts of physical objects to generate and exchange information, e.g., data sensing and transmission by wireless sensors. A critical concern with modern IoT systems is to efficiently utilize limited radio spectrum resources as energy for data transmission. Recently, radio frequency (RF) powered cognitive radio network (CRN) technology has addressed the concern by allowing energy-constrained IoT system devices to recycle energy from RF signals and transmit data using dynamically allocated communication channels~\cite{lee2013opportunistic}. In an RF powered CRN, secondary transmitters (STs) harvest energy from ambient and dedicated RF sources, e.g., RF signals when primary transmitters (PTs) are in transmission. With the harvested energy, secondary transmitters can transmit data to secondary receivers (SRs) using idle primary channels.

However, performance of conventional RF powered CRNs significantly relies on the activities of PTs. When a channel is occupied by a PT, STs cannot transmit data via the occupied channel to avoid collisions among primary and secondary transmissions, which leads to a low throughput of the secondary transmissions if PTs transmit for a long time. To tackle with the channel resources competition among primary and secondary transmissions, and improve the primary channel efficiency, backscatter communication has been applied to allow simultaneous primary and secondary transmissions in CRN systems. In an RF powered backscatter CRN system, an ST can receive, modulate and reflect RF signals from PTs in the presence of ongoing primary transmissions. STs in the system can switch between conventional RF and backscatter communication modes~\cite{lu2018ambient, lu2018wireless}.

An RF powered backscatter ST operates in the following steps. STs encode the transmission signals by specifically designed modulation approaches, and perform secondary transmission along with primary RF signals simultaneously. As the secondary RF signals are essentially the same as the primary RF signals, only with different modulations, backscatter does not introduce severe interference to the primary transmissions~\cite{ruttik2018does}. For example, by adjusting transmission rates and antenna modes, an ST can reflect secondary transmission data using on-off keying (OOK) or frequency-shift keying (FSK)~\cite{choi2015backscatter} modulations. As the backscatter process only involves RF signals receiving and reflecting, power consumption during the backscatter can be low. As a result, backscatter can be a more practical RF based transmission approach compared with conventional RF powered CRN communications. The numerical study in~\cite{hoang2017ambient} shows that the integration of backscatter into RF powered CRN systems always outperforms either conventional CRNs or backscatter systems alone in terms of transmission rate.

We study the CRN system performance when an ST in the system have different transmission behaviors, as follows: (i) {\em Backscatter} mode, where the ST employs backscatter to transmit data; (ii) {\em harvest-then-transmit (HTT)} mode, where the ST harvests energy from primary transmission RF signals and stores the energy for further secondary transmission; and (iii) {\em transmission} mode, where the ST transmits data via idle channels. Optimal behavior decisions have to be made for the ST for optimizing the CRN system performance. For example, problems of balancing between backscatter transmission and energy harvesting (i.e., HTT) have been studied in~\cite{thai2016tradeoff} and \cite{hoang2017optimal}. The objective function to optimize the CRN throughput is formulated as a concave function, where a globally optimal solution can be obtained, i.e., the optimal backscatter/HTT action scheme. STs in~\cite{kim2017hybrid} select between ambient or dedicated RF sources in accordance with locations and environment factors. Throughput of backscatter communications is maximized. An auction approach has been proposed in~\cite{gaobackscatter2019} for assigning backscattering time as a resource. As a classical optimal decision making technique, a Markov Decision Process (MDP) model has been established as in~\cite{van2018reinforcement}, where internal and environment states, e.g., the data queue length, are observed for STs to make backscatter/HTT/transmission decisions to maximize the secondary transmission throughput.

MDP has the drawback to iterate through all system states and update the actions accordingly so as to derive the optimal backscatter/HTT/transmission behavior decisions~\cite{wen2019throughput}, in terms of a maximized long-term reward. However, in large-scaled IoT systems supported by RF powered backscatter CRN, the state space to formulate an MDP model can be extremely large. There can also be unknown or infinite system states, e.g., channel state, which cannot be included in conventional MDP formulations. Consequently, MDP cannot model the CRN system in the case of uncertain and large-scale state space. To address the aforementioned issues, deep reinforcement learning (DRL) approach is applied in this work~\cite{wang2016dueling}, where neural networks are employed to record and learn from the system state and decision records. Optimal decisions for secondary backscatter/HTT/transmission are predicted. In a complex IoT system supported by CRN, the application of DRL is supposed to optimize the secondary transmitter actions with an accelerated convergence and accurate learning process.

\section{System Model}
\label{sec:sys_model}

\begin{figure}
    \centering
    \includegraphics[scale=0.5]{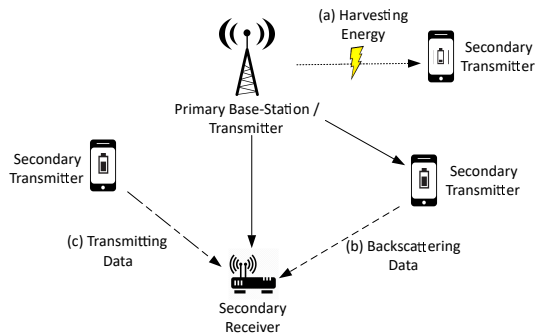}
		\caption{RF-powered Cognitive Radio Network with Various Communication Modes} 
		\label{fig:System_Model}
		\vspace*{-2mm}
\end{figure}

The system model of an RF-powered Backscatter Cognitive Radio Network (CRN) is described in Figure~\ref{fig:System_Model}. The network comprises of a Primary Transmitter (PT), Secondary Transmitter (ST) and Secondary Receiver (SR). PT is modelled to broadcast RF signals on licensed wireless bands, such as Frequency Modulation (FM), Amplitude Modulation (AM) and TV broadcasting Base Station (BS). Within the network, ST is able to operate in three modes: \textit{energy harvesting}, \textit{backscatter} and \textit{active}. Using the onboard energy harvesting circuitry, \textit{energy harvesting} is assumed to occur when the battery level of ST is below 50$\%$ and the energy is stored in an onboard energy storage, such as super capacitor. Data packet transmission may occur during either \textit{backscatter} or \textit{active} modes. ST can transmit signal to SR using \textit{backscatter} mode while the primary channel is busy. Similarly, ST can transmit signal to SR when the primary channel is idle.

\begin{figure}
   \centering   
	\includegraphics[scale=0.5]{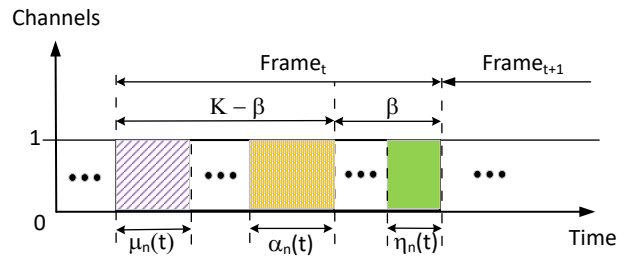} 
	\caption{Time-scheduling of wireless activities} 
	\label{fig:wireless_activity_time_scheduling}
	\vspace*{-2mm}
\end{figure}

For easier understanding, the broadcast signal is presented as a series of time-slots with a fixed duration. For easy understanding, a single channel is assumed and a time-frame comprises of $K$ time-slots, see Figure~\ref{fig:wireless_activity_time_scheduling}. 

Within each time-frame, the number of time-slots for idle ($\beta$) and busy period ($K - \beta$) is random. ST performs data transmission to SR either during busy or idle time-slots. For example, time slots can only be allocated for either backscatter $\alpha_{n}(t)$ or energy harvesting $\mu_{n}(t)$ during the channel busy period. When an individual ST has data for transmission, it will initiate \textit{backscatter} mode. For the remaining channel busy period, \textit{energy harvesting} mode will be initiated. During an idle period ($\beta$), ST is allowed to perform active data transmission to SR, i.e \textit{active} mode operation. $\eta_{n}(t)$ denotes the number of time-slots for $N$ number of ST to remain in \textit{active} mode. Similarly, SR is able to observe the environment and will control the transmission scheduling of the STs in the network. 
\vspace*{-3mm}
\section{Problem Formulation}
\label{sec:problem_formulation}
The objective function of the proposed RF-powered Backscatter CRN network, described in Equation~\ref{eqn:reward_function_1}, is cast into the Markov Decision Process (MDP) framework as a stochastic optimization problem.
\begin{equation}
   \label{eqn:reward_function_1}
   {Throughput}_{max} = \sum_{ST=1}^{N}{PacketsTransmitted}
\end{equation}
An MDP is formally described by a set of states within a state space $\mathcal{S}$, an action space $\mathcal{A}$, a probabilistic distribution that describes the environment dynamics $\mathcal{P}$, and a reward function $\mathcal{R}$ to influence the network's behaviour. 

The state space of secondary transmitter $N$ is denoted as:
\begin{equation}
\mathcal{S}_{n}= (q_{n}, c_{n})
\end{equation}
where {$q_{n} \in {0, 1, ..., \mathcal{Q}_{n}}$} represents the number of packets in the data queue, and {$c_{n}\in{0,1,...,C_{n}}$} represents the energy units in the energy storage. To model the channel state, we let the number of busy time slots be denoted by $\mathcal{S}^{c}=b; b\in0,1,...,K$. The resulting state space for the given network is then denoted as a Cartesian product:
\begin{equation}
	\mathcal{S}=\mathcal{S}^{c} \times \prod_{n=1}^{N}\mathcal{S}_{n}
\end{equation}

Next, the action space of the network is defined as:
\begin{equation}
	\mathcal{A}=\left\{ \begin{array}{l}
			(\mu,\alpha_{1}...,\alpha_{N},\eta_{1},...,\eta_{N})| \\
			\mu+\sum_{n=1}^{N} \alpha_{n} \leq b, \mu+\sum_{n=1}^{N}(\alpha_{n}+\eta_{n})\leq K			
			\end{array} \right.
\end{equation}

The constraints are necessary to ensure that the respective backscatter $\mu_{t}$, harvest $\alpha_{n}$ and active transmission $\eta_{n}$ time slots do not violate the busy $K-\beta$ and idle $\beta$ time-slot for the defined time-frame.

Consider the computation of the network's state transition during the busy period, the number of energy units within the ST storage changes from $c_{n}$ to $c_{n}^{(1)}$, see Equation~\ref{eqn:energyunit_transition_busy}, and $n$ queued data packets in ST changes from $q_{n}$ to $q_{n}^{(1)}$, see Equation~\ref{eqn:dataqueue_transition_busy}. To avoid loss of generality, given a busy time slot, $e_{n}^{h}$ is used to indicate the number of energy units that a ST device is able to harvest while the number of transmitted packets, during \textit{backscatter} mode, is indicated as $d_{n}^{b}$.
\begin{equation}
	c_{n}^{(1)}=min(c_{n}+(\overbrace{K-\beta-\alpha_{n}}^{\text{harvest time-slots}})e_{n}^{h},C_{n})
	\label{eqn:energyunit_transition_busy}
\end{equation}
\begin{equation}
	q_{n}^{(1)}=max(0, q_{n}-\alpha_{n}d_{n}^{b})	
	\label{eqn:dataqueue_transition_busy}
\end{equation}

Likewise, consider the condition where the channel is idle. To avoid loss of generality, $e_{n}^{a}$ is used to indicate the number of energy units that a ST device is able to harvest while the number of transmitted packets, during \textit{backscatter} mode, is indicated as $d_{n}^{a}$. Data transmission is possible in $min(\eta_{n}, q_{n}^{(1)}/d_{n}^{a})$ time-slots. Once the idle channel period has elapsed, the energy state of $n$ ST changes from $c_{n}^{(1)}$ to $c_{n}^{(')}$, see Equation~\ref{eqn:energyunit_transition_idle}, while the queued data packets in ST changes from $q_{n}^{(1)}$ to $q_{n}^{(2)}$, see Equation~\ref{eqn:dataqueue_transition_idle}. The expression $q_{n}^{1}/d_{n}^{a}$ describes the time slots required, by $n$ ST, to transmit $q_{n}^{1}$ data packets.
\begin{equation}
	c_{n}^{'}=max[0, c_{n}^{(1)}-min(\eta_{n}, q_{n}^{(1)}/d_{n}^{a}]e_{n}^{a}
	\label{eqn:energyunit_transition_idle}
\end{equation}
\begin{equation}
	q_{n}^{(2)}=max[0, q_{n}^{(1)}-min(\eta_{n}, c_{n}^{(1)}/e_{n}^{a}]d_{n}^{a}
	\label{eqn:dataqueue_transition_idle}
\end{equation}
To simulate the packet arrival of new data packets within $K$ time-slots, a binomial distribution behaviour $B(K, \lambda_{n})$ is assumed, where $\lambda_{n}$ is the probability of new packets arriving at each time-slot.
\begin{equation}
	\mathcal{P}(p_{n}=m)=\left({\begin{array}{*{10}{c}}
	F\\
	m
	\end{array}}\right)\lambda_{n}^{m}(q-\lambda_{n})^{K-m}
\end{equation}

Designing the reward function to optimize time-scheduling of STs, from SR perspective, requires the reward function $\mathcal{R}$ to be a function of the state $\mathcal{S}$ and actions $\mathcal{A}$ of the network. With the previous derivations, $\mathcal{R}$ is re-defined as the accumulated number of packets transmitted with respect to operational modes as defined in Equation~\ref{eqn:reward_function}.
\begin{equation}
\label{eqn:reward_function}
	\mathcal{R}(s,a)=\underbrace{\sum_{n=1}^{N} S_{n}^{b}(q_{n}^{(1)}-q_{n})}_{\text{backscatter}}  + \underbrace{\sum_{n=1}^{N} S_{n}^{a}(q_{n}^{(2)}-q_{1})}_{\text{active}} 
\end{equation}

As a result, the optimal policy $(\pi^{*})$ can be obtained by maximizing the value-state function:
\begin{equation}
	\mathcal{V}(s)=\mathbb{E}[\sum_{t=0}^{\mathcal{T}-1}\gamma\mathcal{R}(s_{t}, a{_t})] 
\end{equation}
where $\mathcal{T}$ denotes the time-horizon duration, $\gamma$ is the discount factor for $0\leq \gamma<1$ and expected value(s) $\mathbb{E}[]$. Considering the Markov property, the value function is further expressed as:
\begin{equation}
	\label{eqn:value_func}
	\mathcal{V}(s)=\sum_{s' \in S}\mathcal{P}_{\pi(s)}(s, s')(\mathcal{R}(s, a)+\gamma\mathcal{V}(s'))
\end{equation}
The associated policy function obtains the maximum action that is possible from Equation~\ref{eqn:value_func}. Hence, the Q-function $\mathcal{Q}^{new}(s, a)$ can be updated using the Bellman equation and expressed as:
\begin{equation*}
   \begin{multlined}
      \overbrace{\mathcal{Q}^{new}(s, a)}^{\text{New Q value}}= (1-\alpha)\overbrace{\mathcal{Q}(s, a)}^{\text{Current Q value}}+\alpha[\overbrace{\mathcal{R}(s, a)}^{\text{Reward received}}\\
+\gamma \underbrace{\max\limits_{a' \in \mathcal{A}}\mathcal{Q}'(s', a')}_{\text{Max(Expected future reward)}}]
   \end{multlined}
\end{equation*}

Note that $\alpha$ denotes the learning rate of the Q-network. The discount factor $0\leq\gamma<1$ is typically used to shape the behaviour of the agent by determining the importance of the observed reward. A value of 0 makes the agent place greater importance on immediate or short-term rewards while value close to 1 encourages the agent to place greater emphasis on longer-term reward. 

In Q-learning, the agent selects an action to perform, based on the Q-values stored within a look-up table. The Q-value is iteratively updated and the expected performance of the algorithm decreases exponentially\cite{wen2019throughput} as the observable state and action space becomes intractably large. To overcome the issue, the function approximator method is proposed for estimation of optimal Q-function. Hence,  Artificial Neural Networks becomes a natural candidate to select. 
\vspace*{-1mm}	
\section{Deep Reinforcement Learning}
\label{sec:DQNintro}
The combination of Q-learning with deep neural network is known as Deep Q-Network (DQN) or Deep Reinforcement Learning (DRL). Note that the terms DQN and DRL is used interchangeably. In particular, the deep neural network is used to estimate the Q-values for each state-action pair, for a large environment, before an optimal Q-function can be approximated. In our system model, the inputs to DQL will contain a tuple of randomly generated timeslots for each ST to perform backscattering, energy harvesting and active data transmission. The DQN output includes Q-values $\mathcal{Q}(s,a;\theta)$ for all possible actions of the Secondary Transmitter; $\theta$ represents the weights of the deep neural network for the derivation of the next state's Q-value. At the end of every episode, the max operator helps identify the best possible action of the gateway which enables it to obtain the best possible reward, which is then stored in the replay memory buffer. The network's loss value $\Delta w$ is defined as the difference in the target reward value and current reward value. The loss value is then back-propagated throughout the deep neural network to update its weights $\theta$ for minimizing the loss function.
\begin{equation} 
\label{eqn:loss_weights}
	\Delta w = \alpha[TD_{error}]\nabla_{Q}
\end{equation}
where the gradient of our current predicted Q-value($\nabla_{Q}$) is:
\begin{equation}
	\nabla_{Q} = \nabla_{w} \mathcal{\hat{Q}} (s, a, w)
\end{equation}

The network's learning rate $\alpha$ is a hyper-parameter that controls the rate of updating the network weights with respect to the loss gradient value. $TD_{error}$ is calculated by taking the difference between the Q-target (maximum possible value from next state) and Q-value (our current prediction of the Q-value). The mathematical representation is denoted in Equation~\ref{eqn:TD_error}:
\begin{equation}
\label{eqn:TD_error}
	TD_{error}=\overbrace{{\mathcal{R} + \gamma max_{a} \mathcal{\hat{Q}}(s', a, \theta^{-}))}}^{\text{Max Q-value for next state (Q-target)}}-\overbrace{\mathcal{\hat{Q}}(s, a, w)}^{\text{Predicted Q-value}}
\end{equation}
To simplify, the target Q-value is denoted as $y$ follows where $\theta^{-}$ represents the weights from the previous iteration, see Equation~\ref{eqn:y_DQN}.
\vspace*{-5mm}
\begin{equation}
\label{eqn:y_DQN}
y=r + \gamma max \mathcal{Q}(s', a', \theta^{-})
\end{equation}
The $\epsilon$-greedy algorithm is a technique to constantly stimulate the DQN agent to perform exploration whilst picking actions which, known to perform well. For example, given the current value of the $\epsilon$-greedy policy, the DQN agent has a probability to either explore the environment and select a random action, or exploit the environment and select the greedy action i.e maximum Q-value or reward. As the $\epsilon$ value approaches $0$, the DQN agent switches to greedy policy and will instead start exploiting its accumulated knowledge i.e the experience replay buffer. The pseudocode for the proposed DQN algorithm, for the wireless gateway, is described in Algorithm~\ref{algo:DQN_ReplayTarget} below. 
\begin{algorithm}[!htbp]
\caption{Deep Q-Learning with Experience Replay for Gateway Time-scheduling}
\begin{algorithmic}[1]
\State \textbf{Input:} Action space $\mathcal{A}$, mini-batch size $L_{b}$, target network replacement frequency $L^{-}$
\State \textbf{Output:} Optimal policy $\pi^{*}$ for $N$ Secondary Transmitters

\State Initialize replay memory $\mathcal{D}$ to capacity $N$
\State Initialize action-value function $\mathcal{Q}$ with random weights
\State Initialize target action-value function $\mathcal{\hat{Q}}$ with weights $\mathcal{\theta^{-}}$=$\mathcal{\theta}$
\For{Episode=1 to $E$}
	\State Initialize sequence $s_{1}={x_{1}}$ and preprocessed sequence $\Phi_{1}=\Phi(s_{1})$
	\For{timestep=1 to T}
		\State Choose an action $a_{t}$ 
		\State    With probability $\mathcal{\epsilon}$, a random action is performed
		\State    Otherwise, choose $a_{t}=argmax_{a}\mathcal{Q}(\Phi(s_{t}, a)$ from $\mathcal{Q}(s,a;\theta)$
		\State Broadcast messaging time-schedules for $N$ secondary transmitters
		\State Execute chosen action $a$ 
		\State Receive reward $r$
		\State Receive state messages from primary transmitter and $N$ secondary Transmitters
		\State Update next network state $s'$
		\State Store tuple $(s, a, r, s')$ in replay memory $D$
		\State Randomly sample tuple $(ss, aa, r{r}, ss')$ of mini-batch size $(L_{b})$ from replay memory $\mathcal{D}$ 
		\State Calculate target Q-value for each mini-batch transition
		\State $y_{t}^{DQN}=\left\{ \begin{array}{l}
			r, \text{if episode i terminates at timestep+1} \\
			r+\gamma max_{a'}\mathcal{\hat{Q}}(\mathcal{\phi}_{j+1}, a', \mathcal{\theta}^{-}),\text{else}
			\end{array} \right.$
		\State Train the Q-Network using $(y_{t}^{DQN}-\mathcal{Q}(ss, aa)^{2})$ as loss and update the weights $\theta$
		\State Reset $\theta^{-}$=$\theta$ every $L^{-}$ steps
		\State Update $s \leftarrow s'$ 
		\State Increment timestep by 1
	\EndFor \textbf{repeat until} timestep is > T, terminate
\EndFor \textbf{repeat until} Episode is > $E$, terminate
\end{algorithmic}
\label{algo:DQN_ReplayTarget}
\end{algorithm}
\vspace*{-3.5mm}
\section{Performance Evaluation and Results}
\label{sec:performance_results}
\begin{table}
	\centering
	\vspace*{-1mm}
	\begin{tabular}{cc} 
		 & \\ \hline
		\textbf{Parameter} & \textbf{Value} \\ \hline \hline
		Hidden Layers & 1(DQN), 3(Comparison) \\ \hline
		Number of Hidden Neurons ($H_{n}$) & 16, 32 ,64, 128, 256 \\ \hline
		Optimizer & Adam, SGD \\ \hline
		$\epsilon$-Greedy decay & 0.9$\rightarrow$0 \\ \hline
		$\epsilon$-Greedy decay steps & $4\times10^{5}$ \\ \hline
		Learning Rate ($\alpha$) & $1e^{-3}$, $1e^{-4}$ \\ \hline
		Discount rate ($\gamma$) & 0.9 \\ \hline
		Target Network Update Rate & $1e^{-4}$ \\ \hline
		Mini-batch size & 32 \\ \hline
		Replay Memory size & $5\times10^{5}$ \\ \hline
		Iteration steps per Episode & 200 \\ \hline
		Training iterations & $10^{6}$ \\ \hline
		Secondary Transmitters (N) & 2,3 \\ \hline
		Time slots within single time frame & 10 \\ \hline
		Idle time slots within single time frame & [1;9] \\ \hline
		Packet Arrival Probability ($\lambda_{n}$) & [0.1;0.9] \\ \hline
	\end{tabular}	
	\caption{DQN Model Simulation Parameters}
	\label{tab:tbl_hyperparameter}
\end{table}

\begin{table}[htb]
	\centering
	\begin{tabular}{|c|c|c|c|c|} \hline 
		\textbf{Environment} & \textbf{Number of Neurons} & \textbf{Adam} & \textbf{SGD} & \textbf{Speedup}\\ \hline
		2ST & 16 & 183 & 269 & \textasciitilde{}1.5x \\ \hline
		2ST & 32 & 210 & 379 & \textasciitilde{}1.8x \\ \hline
		2ST & 64 & 212 & 203 & \textasciitilde{}0.96x \\ \hline
		2ST & 128 & 246 & 288 & \textasciitilde{}1.2x \\ \hline
		2ST & 256 & 184 & 283 & \textasciitilde{}1.5x \\ \hline	\hline
		3ST & 16 & 1794 & 1631 & \textasciitilde{}0.91x \\ \hline
		3ST & 32 & 1792 & 1675 & \textasciitilde{}0.93x \\ \hline
		3ST & 64 & 1792 & 1561 & \textasciitilde{}0.87x \\ \hline
		3ST & 128 & 1792 & 1571 & \textasciitilde{}0.88x \\ \hline
		3ST & 256 & 1763 & 1678 & \textasciitilde{}0.95x \\ \hline
	\end{tabular}
	\caption{Optimizer Speedup for varying Number of Hidden Neurons (Single Hidden Layer)}		  
	\label{tab:Optimizer_Compare}
	\vspace*{-3mm}
\end{table} 

\begin{table*}[htb]
	\centering
	\caption{Mean Training performance for various DQN techniques}
	\label{tab:DQNTechniques_Comparison}
	\begin{tabularx}{\textwidth}{|c|m{2.4cm}|m{1.8cm}|c|c|m{3.3cm}|c|} \hline 
		\textbf{Environment} & \centering \textbf{DQN Method} & \centering \textbf{Optimizer} & \textbf{Hidden Neurons} & \textbf{Layers} & \centering \textbf{Mean Throughput(pkts)} & \textbf{Speedup wrt DoubleDQN} \\ \hline
		2ST &  \centering DQN-SGD32 & \centering SGD & 32 & 1 & \centering 379 & \textasciitilde{3.1x} \\ \hline
		2ST &  \centering DQN-Adam128 & \centering Adam & 128 & 1 & \centering 246 & \textasciitilde{2.0x} \\ \hline
		2ST &  \centering DoubleDQN & \centering Adam & 32 & 3 & \centering 124 & NA \\ \hline
		2ST &  \centering DuelDQN & \centering Adam & 32 & 3 & \centering 224 & \textasciitilde{1.8x} \\ \hline
		2ST &  \centering DoubleDuelDQN & \centering Adam & 32 & 3 & \centering 173 & \textasciitilde{1.4x} \\ \hline \hline
		3ST &  \centering DQN-SGD32 & \centering SGD & 32 & 1 & \centering 1675 & \textasciitilde{1.07x} \\ \hline
		3ST &  \centering DQN-Adam128 & \centering Adam & 128 & 1 & \centering 1793 & \textasciitilde{1.15x} \\ \hline
		3ST &  \centering DoubleDQN & \centering Adam & 32 & 3 & \centering 1560 & NA \\ \hline
		3ST &  \centering DuelDQN & \centering Adam & 32 & 3 & \centering 1731 & \textasciitilde{1.11x} \\ \hline
		3ST &  \centering DoubleDuelDQN & \centering Adam & 32 & 3 & \centering 1767 & \textasciitilde{1.13x} \\ \hline
	\end{tabularx}
	\vspace*{-3mm}	
\end{table*} 
\subsection{Parameter Settings}
The DQN techniques~\cite{mnih2013playing, wang2016dueling, van2016deep, anh2018deep} were implemented in Tensorflow. To optimize the agent's performance, both $\epsilon$-Greedy algorithm and replay memory tweaks were utilized. A fully-connected Deep Neural Network (DNN) architecture is proposed and the hyperparameter configurations are detailed in Table~\ref{tab:tbl_hyperparameter}. 

For each simulation scenario, the reward function is implemented as described in Equation~\ref{eqn:reward_function_1}, from the wireless gateway perspective. The actions to be performed are defined as the time frame assignment of each secondary transmitters to perform backscatter, harvest-then-transmit (HTT) and transmit data. Optimal policy is assumed to be learnt when the agent's maximum reward stabilizes for $\geq$100 episodes. Parameter details of the simulation environments can be found in Table~\ref{tab:tbl_hyperparameter}. For analysis and readability purposes, reported results reflect the mean values for 10 runs. 
\vspace*{-3.5mm}
\subsection{Results}

The number of neurons $H_{n}$, within the hidden layer, dictates the learning capacity of DQN algorithm and unless the inflexion point is reached, an increase in neurons should lead to improved network throughput performance. Considering the Adam results, the assumption was only valid for environment with 2-STs with the maximum throughput occurring when $H_{n}$=128. Similarly, the inflexion point occurred much earlier at $H_{n}$=16, for environment with 3-STs. Performance remained constant before further degradation at $H_{n}$=256. For further details, readers are referred to Table~\ref{tab:Optimizer_Compare}. 



Stochastic Gradient Descent (SGD) was also tested to provide a different perspective of the gradient landscape and notable performance difference between SGD and Adaptive Moment estimation (Adam) optimizers has been observed. To quantify the performance gains or degradation of SGD performance with respect to the Adam optimizer performance, the Speedup metric is introduced. Additional details can be found in Table~\ref{tab:Optimizer_Compare}.


The best performing configurations of SGD optimizer (SGD-32) and Adam optimizer (Adam-128) were benchmarked against advanced DQN techniques reported in~\cite{anh2018deep}. Although our proposed DQN network configuration is lightweight, it provided performance speedup of between 1.07x to 3.1x, with all techniques benchmarked against the slowest performer - DoubleDQN. For convenience, the performance summary is illustrated in Table~\ref{tab:DQNTechniques_Comparison}. 
\subsection{Analysis and Future Research}
We have provided empirical proof that our proposed lightweight DQN configurations, SGD-32 and ADAM-128, outperformed several advanced DQN techniques. Considering the DQN architecture alone, the optimizer selection has shown strong correlation to the DQN agent's training performance and similar conclusion was reported in \cite{wilson2017marginal}. We had also observed that a reduction in the $\epsilon$-greedy steps reduced training time by as much as 50$\%$ with minor performance degradation. Next, the intractable nature of performing a full grid hyperparameter search meant that ADAM's learning rate was defaulted to \cite{anh2018deep} and only learning rate tuning was performed to ensure optimal solution convergence, using SGD, within given simulation time-steps.

Due to time and space constraints, the provided use-case was simplified. Future research work could include extended discussions on the performance scaling, for increasing STs, multiple PUs and STs and multi-channel scenario.  

\section{Conclusions}
\label{sec:conclusion}
In this paper, time-scheduling of a wireless secondary receiver, given a complex network environment, has been formulated into a stochastic optimization problem. The proposed DQN algorithm was able to derive an optimal policy within 2000 episodes. In comparison with several advanced DQN techniques, our proposed lightweight DQN configuration is able to learn an optimal time-scheduling policy with an overall network throughput performance speedup of up to 1.8x. 
\bibliographystyle{IEEEtran}
\bibliography{Backscatterdatabase_clean}

\begin{thebibliography}{10}
\providecommand{\url}[1]{#1}
\csname url@samestyle\endcsname
\providecommand{\newblock}{\relax}
\providecommand{\bibinfo}[2]{#2}
\providecommand{\BIBentrySTDinterwordspacing}{\spaceskip=0pt\relax}
\providecommand{\BIBentryALTinterwordstretchfactor}{4}
\providecommand{\BIBentryALTinterwordspacing}{\spaceskip=\fontdimen2\font plus
\BIBentryALTinterwordstretchfactor\fontdimen3\font minus
  \fontdimen4\font\relax}
\providecommand{\BIBforeignlanguage}[2]{{%
\expandafter\ifx\csname l@#1\endcsname\relax
\typeout{** WARNING: IEEEtran.bst: No hyphenation pattern has been}%
\typeout{** loaded for the language `#1'. Using the pattern for}%
\typeout{** the default language instead.}%
\else
\language=\csname l@#1\endcsname
\fi
#2}}
\providecommand{\BIBdecl}{\relax}
\BIBdecl

\bibitem{lee2013opportunistic}
S.~Lee, R.~Zhang, and K.~Huang, ``Opportunistic wireless energy harvesting in
  cognitive radio networks,'' \emph{IEEE Transactions on Wireless
  Communications}, vol.~12, no.~9, pp. 4788--4799, 2013.

\bibitem{lu2018ambient}
X.~Lu, D.~Niyato, H.~Jiang, D.~I. Kim, Y.~Xiao, and Z.~Han, ``Ambient
  backscatter assisted wireless powered communications,'' \emph{IEEE Wireless
  Communications}, vol.~25, no.~2, pp. 170--177, 2018.

\bibitem{lu2018wireless}
X.~Lu, H.~Jiang, D.~Niyato, D.~I. Kim, and Z.~Han, ``Wireless-powered
  device-to-device communications with ambient backscattering: Performance
  modeling and analysis,'' \emph{IEEE Transactions on Wireless Communications},
  vol.~17, no.~3, pp. 1528--1544, 2018.

\bibitem{ruttik2018does}
K.~Ruttik, R.~Duan, R.~J{\"a}ntti, and Z.~Han, ``Does ambient backscatter
  communication need additional regulations?'' in \emph{2018 IEEE International
  Symposium on Dynamic Spectrum Access Networks (DySPAN)}.\hskip 1em plus 0.5em
  minus 0.4em\relax IEEE, 2018, pp. 1--6.

\bibitem{choi2015backscatter}
S.~H. Choi and D.~I. Kim, ``Backscatter radio communication for wireless
  powered communication networks,'' in \emph{2015 21st Asia-Pacific Conference
  on Communications (APCC)}.\hskip 1em plus 0.5em minus 0.4em\relax IEEE, 2015,
  pp. 370--374.

\bibitem{hoang2017ambient}
D.~T. Hoang, D.~Niyato, P.~Wang, D.~I. Kim, and Z.~Han, ``Ambient backscatter:
  {A} new approach to improve network performance for {RF}-powered cognitive
  radio networks,'' \emph{IEEE Transactions on Communications}, vol.~65, no.~9,
  pp. 3659--3674, 2017.

\bibitem{thai2016tradeoff}
H.~D. Thai, D.~Niyato, P.~Wang, D.~I. Kim, and Z.~Han, ``The tradeoff analysis
  in {RF}-powered backscatter cognitive radio networks,'' in \emph{2016 IEEE
  Global Communications Conference (GLOBECOM)}.\hskip 1em plus 0.5em minus
  0.4em\relax IEEE, 2016, pp. 1--6.

\bibitem{hoang2017optimal}
D.~T. Hoang, D.~Niyato, P.~Wang, and D.~I. Kim, ``Optimal time sharing in
  {RF}-powered backscatter cognitive radio networks,'' in \emph{2017 IEEE
  International Conference on Communications (ICC)}.\hskip 1em plus 0.5em minus
  0.4em\relax IEEE, 2017, pp. 1--6.

\bibitem{kim2017hybrid}
S.~H. Kim and D.~I. Kim, ``Hybrid backscatter communication for
  wireless-powered heterogeneous networks,'' \emph{IEEE Transactions on
  Wireless Communications}, vol.~16, no.~10, pp. 6557--6570, 2017.

\bibitem{gaobackscatter2019}
X.~{Gao}, P.~{Wang}, D.~{Niyato}, K.~{Yang}, and J.~{An}, ``Auction-based time
  scheduling for backscatter-aided {RF}-powered cognitive radio network,''
  \emph{IEEE Transactions on Wireless Communications}, Early Access, 2019.

\bibitem{van2018reinforcement}
N.~Van~Huynh, D.~T. Hoang, D.~N. Nguyen, E.~Dutkiewicz, D.~Niyato, and P.~Wang,
  ``Reinforcement learning approach for {RF}-powered cognitive radio network
  with ambient backscatter,'' \emph{arXiv preprint arXiv:1808.07601}, 2018.

\bibitem{wen2019throughput}
X.~Wen, S.~Bi, X.~Lin, L.~Yuan, and J.~Wang, ``Throughput maximization for
  ambient backscatter communication: {A} reinforcement learning approach,''
  \emph{arXiv preprint arXiv:1901.00608}, 2019.

\bibitem{wang2016dueling}
Z.~Wang, T.~Schaul, M.~Hessel, H.~Hasselt, M.~Lanctot, and N.~Freitas,
  ``Dueling network architectures for deep reinforcement learning,'' in
  \emph{International Conference on Machine Learning}, 2016, pp. 1995--2003.

\bibitem{mnih2013playing}
V.~Mnih, K.~Kavukcuoglu, D.~Silver, A.~Graves, I.~Antonoglou, D.~Wierstra, and
  M.~Riedmiller, ``Playing {Atari} with deep reinforcement learning,''
  \emph{arXiv preprint arXiv:1312.5602}, 2013.

\bibitem{van2016deep}
H.~Van~Hasselt, A.~Guez, and D.~Silver, ``Deep reinforcement learning with
  double {Q}-learning,'' in \emph{Thirtieth AAAI Conference on Artificial
  Intelligence}, 2016.

\bibitem{anh2018deep}
T.~T. Anh, N.~C. Luong, D.~Niyato, Y.-C. Liang, and D.~I. Kim, ``Deep
  reinforcement learning for time scheduling in {RF}-powered backscatter
  cognitive radio networks,'' \emph{arXiv preprint arXiv:1810.04520}, 2018.

\bibitem{wilson2017marginal}
A.~C. Wilson, R.~Roelofs, M.~Stern, N.~Srebro, and B.~Recht, ``The marginal
  value of adaptive gradient methods in machine learning,'' in \emph{Advances
  in Neural Information Processing Systems}, 2017, pp. 4148--4158.

\end{thebibliography}

\end{document}